\documentclass{article}
\usepackage[final,nonatbib]{nips_2016}

\usepackage[utf8]{inputenc} 
\usepackage[T1]{fontenc}    
\usepackage{hyperref}       
\usepackage{url}            
\usepackage{booktabs}       
\usepackage{amsfonts}       
\usepackage{nicefrac}       
\usepackage{microtype}      

\usepackage{graphicx} 
\usepackage{amsmath} 
\usepackage{amssymb}
\usepackage{verbatim}

\renewcommand\vec{\mathbf}

\title{Real-time interactive sequence generation and control \\ with Recurrent Neural Network ensembles}

\author{
	Memo Akten\\
	Department of Computing\\
	Goldsmiths University of London\\
	\texttt{m.akten@gold.ac.uk} \\
	\And
	Mick Grierson\\
	Department of Computing \\
	Goldsmiths University of London \\
	\texttt{m.grierson@gold.ac.uk}	
}

\begin{document}

\maketitle

\begin{abstract}
Recurrent Neural Networks (RNN), particularly Long Short Term Memory (LSTM) RNNs, are a popular and very successful method for learning and generating sequences. However, current generative RNN techniques do not allow real-time interactive control of the sequence generation process, thus aren't well suited for \textit{live creative expression}. We propose a method of real-time continuous control and `steering' of sequence generation using an ensemble of RNNs and dynamically altering the mixture weights of the models. We demonstrate the method using character based LSTM networks and a gestural interface allowing users to `conduct' the generation of text.
\end{abstract}

\section{Introduction}
\textit{Recurrent Neural Networks (RNN)} are artificial neural networks with recurrent connections, allowing them to learn temporal regularities and model sequences. \textit{Long Short Term Memory (LSTM)} \cite{Hochreiter1997} is a recurrent architecture that overcomes the problem of gradients exponentially vanishing \cite{Hochreiter1991,bengio1994learning}, and allows RNNs to be trained many time-steps into the past, to learn more complex programs \cite{Schmidhuber2015}. Now, with increased compute power and large training sets, LSTMs and related architectures are proving successful not only in sequence classification \cite{Graves2009,Hinton2012,Pham2014,Greff2015}, but also in sequence generation in many domains such as music \cite{Eck2002,Boulanger-Lewandowski2012,Nayebi2015,Sturm2015}, text \cite{Martens2011,Sutskever2013}, handwriting \cite{Graves2013}, images \cite{Gregor2015}, machine translation \cite{SutskeverVinyals2014}, speech synthesis \cite{Wu2016} and even choreography \cite{friis2016}.

However, most current applications of sequence generation with RNNs is not a real-time, interactive process. Some recent implementations have used a turn-based approach, such as the online text editor Word Synth \cite{goodwin-wordsynth}. This allows a user to enter a `seed' phrase for the RNN, `priming' it such that the next phrase generated is conditioned on the seed. Although a very useful approach, this still does not provide real-time continuous control in the manner required for the creation of expressive interfaces.

\begin{figure*}
	\centering
	\includegraphics[width=\linewidth]{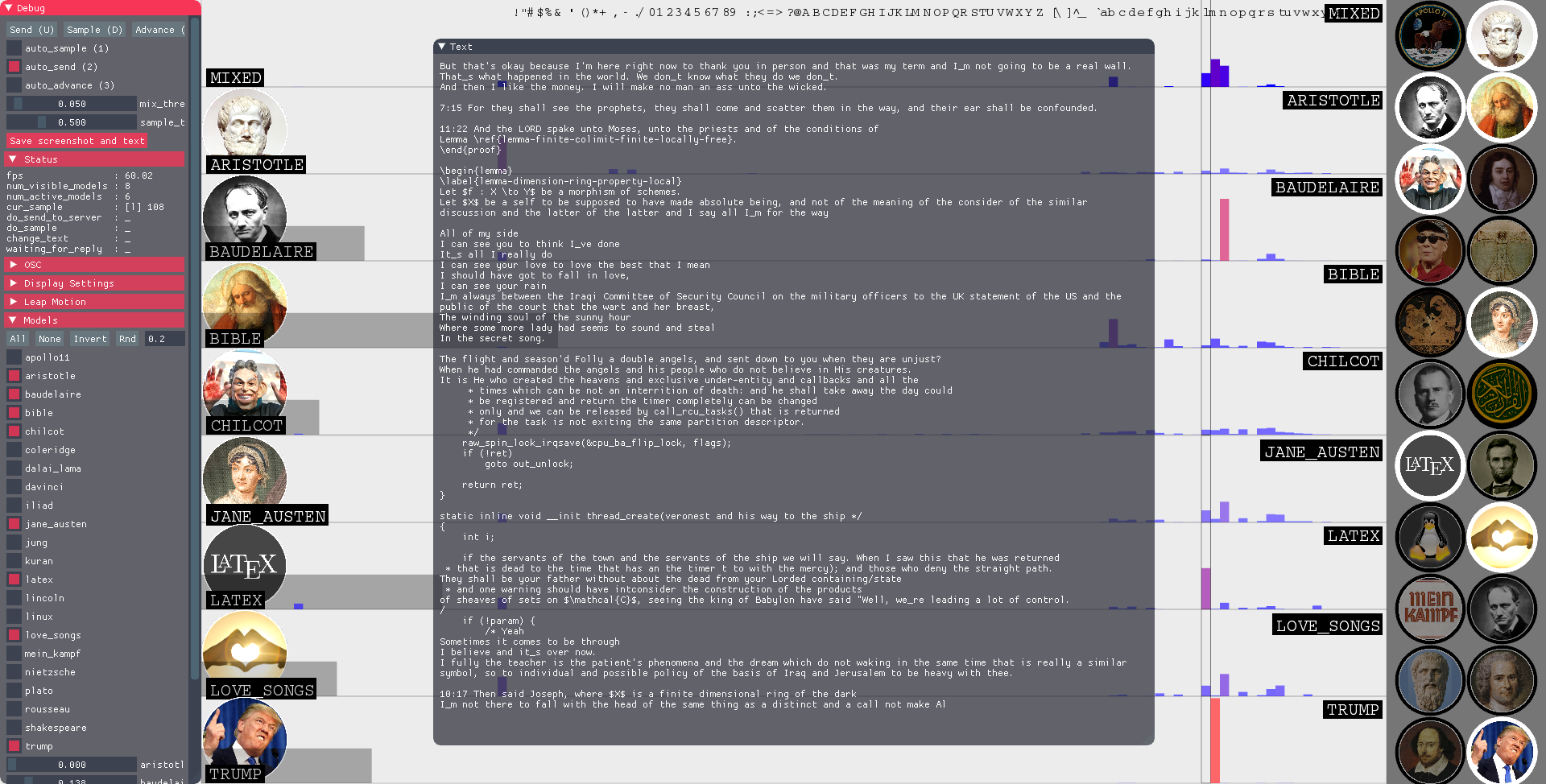}
	\caption{Example screen output and generated text. The user can select multiple models with over 20 to choose from. The chosen models are instantly available to interact with. The blue-red vertical bars visualise the probability distributions at the current time-step $t$, one row per active model ($\vec{y}_t^i$ for the $i$th model), and the joint probability distribution $\vec{\rho}_t$ (the top row). The horizontal dark grey bars on each row visualise the mixture weights for each model, stored in the vector $\vec{\pi}_t$.}
	\label{fig:screen}
\end{figure*}


\section{Method}
An \textit{ensemble} of models is usually used to improve overall prediction accuracy. The common motivation behind this approach is that training multiple diverse models (using different architectures, parameters and/or algorithms) and then combining their predictions (through weighted or unweighted `voting' or averaging) is likely to minimise bias and undesired variance, and thus is more likely to provide more accurate results \cite{Dietterich2000}. Usually in these cases, all models are trained on the same training data. 

We propose a method of using an RNN ensemble, containing models trained on \textit{vastly different} datasets, and dynamically altering the models' mixture weights in real-time to control the output `style' \footnote{We use the term 'style' very liberally here.}. While this method can potentially be applied to many different domains, we choose to first demonstrate it on character based text models for a number of practical reasons: i) the data is relatively low dimensional and has modest computational requirements (processing power, memory requirements, training times etc.), ii) training data is very easy to find, iii) the output is simple to judge qualitiatively and unambigously, iv) it has been demonstrated that LSTMs are successful in this domain \cite{Graves2013,Karpathy2015a}.

\subsection{Training data}
We train an ensemble consisting of $n$ LSTM networks, each trained on a different corpus of text representing a unique style. The styles were selected due to the ease with which each can be characterised with respect to language use and structure. They include the works of Shakespeare, Baudelaire, Nietzsche, Jane Austen, Donald Trump speeches, the King James Bible, assorted love song lyrics, Linux kernel C code, \LaTeX{} source, the Chilcot Report of the Iraq Inquiry and many more. The amount of training data varies for each corpus, ranging from 500KB to 10MB.

\subsection{Training}
We use different architectures for each model depending on the size of the training data, ranging from a single LSTM layer with 256 dimensions, to three LSTM layers each with 512 dimensions. We use LSTM cells with input, output and forget gates \cite{Gers2000}, without peepholes or skip connections between layers. We use Dropout regularization as described in \cite{Zaremba2014} with a dropout probability ranging from 10\% to 30\% depending on the model and architecture. 

In order to provide cross-model compatibility of inputs and outputs, we use a consistent mapping between characters and indices. So we choose standard ASCII codes with each model having input and output dimensions of 128, with a softmax on the output to provide a probability distribution over the 128 characters. We train each model to minimize the negative log-likelihood of the next character given a sequence of characters (of maximum length 80), as described in \cite{Graves2013}.

\begin{figure*}
	\centering
	\includegraphics[width=\linewidth]{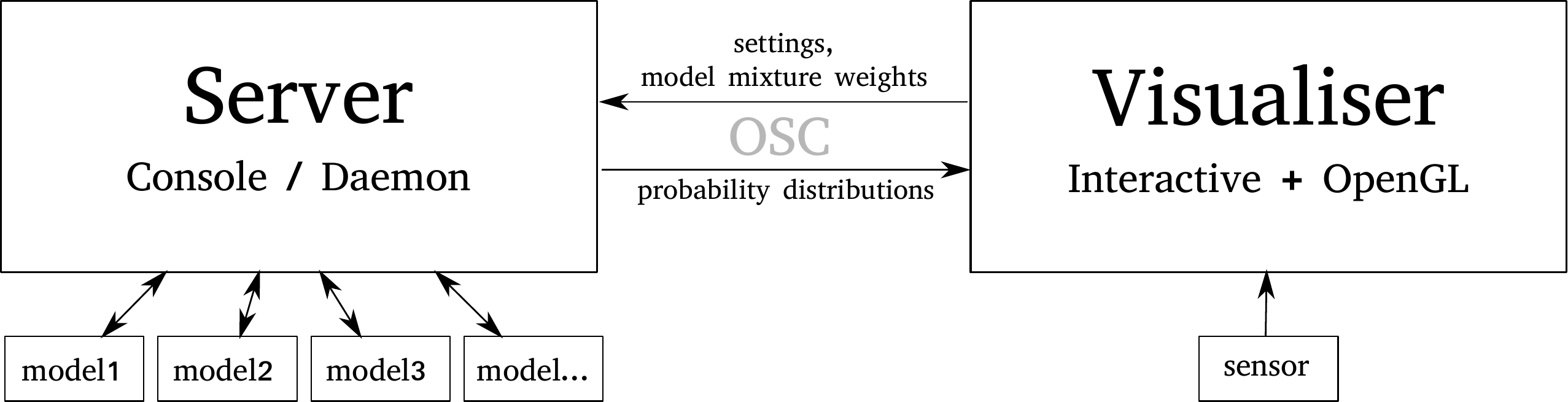}
	\caption[Prediction software architecture]{Software architecture for the interactive prediction and visualisation system. The \textit{Visualiser} is an OpenGL application which continually calculates model mixture weights $\vec{\pi}_t$ at each time-step $t$, either via mouse input, tracking the user's hands using a LeapMotion device or using an external midi controller. The Visualiser sends $\vec{\pi}_t$ via the OSC protocol to the \textit{Server}, which runs each of the models independently on the same input $\vec{x}_t$ to receive probability distributions from each model $\vec{y}_t^i$ for the next character. The Server then sends each $\vec{y}_t^i$ back to the Vizualizer which calculates the joint $\vec{\rho}_t$ weighted by $\vec{\pi}_t$. Separating the two processes allows transparently switching between running both processes on the same computer, or running the backend on a remote, more powerful server.}
	\label{fig:rnn_prediction_sw_arch}
\end{figure*}

\subsection{Interactive prediction and visualisation}
Once trained, the system loads and runs each of the models independently with the same character input represented as a one-hot vector $\vec{x}_t$. Each model --- parameterized by $\vec{\theta}^i$ where the superscript $i$ denotes the model index $\{i \in [1, n]\}$ --- predicts a probability distribution for the next character conditioned on the current history of inputs and can be written as 
\begin{equation}
\vec{y}_t^i = P(\vec{x}_{t+1} | \vec{x}_1, \vec{x}_2,..., \vec{x}_t, \vec{\theta}^i)
\end{equation}

These are stored in a conditional probability matrix $\vec{\Omega}_t$ where the $i$th row contains $\vec{y}_t^i$, the distribution predicted by the $i$th model, i.e. conditioned on the $i$th dataset. The system then calculates the joint distribution at time $t$ by mixing each model's output via a \textit{model mixture weights} vector $\vec{\pi}_t$ --- which can be thought of as the marginal distribution $P(\theta^i)$ --- with

\begin{align}
\vec{\rho}_t &= \frac{\vec{\pi}_t \times \vec{\Omega}_t}{\vert \vec{\pi}_t \times \vec{\Omega}_t \vert_1} \; :  \text{joint distribution} \\
\text{where} \\
\vec{\Omega}_t &: \text{conditional probability matrix. }i \text{th row contains} \; \vec{y}_t^i\\
\vec{\pi}_t &: \text{model mixture weights, i.e. marginal distribution} \; P(\theta^i) \\
\end{align}

where the denominator is simply a normalising factor. Finally, the system samples a character from $\vec{\rho}_t$, prints it to the screen, and feeds it back into the system at the next time-step as $\vec{x}_t$ with $t \leftarrow t+1$.

While the sequence is being generated, a user can steer the output towards different models by interacting with the system and dynamically shifting the mixture weights $\vec{\pi}_t$. Interaction is through clicking on the screen, hand-tracking with a LeapMotion device, or through the use of an external midi controller \footnote{We are also developing new interaction mechanisms such as using a Multilayer Perceptron to map the user's facial expressions or hand gestures to different configurations of mixture weights.}. As an optimization, at every time-step we only run models which have a mixture weight $>5\%$. Depending on the number of models active, the system outputs characters at around 5-20 chars/second on a high-end gaming laptop. 

With this method we are able to guide the system to morph between the different models' output with relatively smooth transitions between different styles. 




\subsection{Software architecture}
The interactive prediction system consists of two standalone processes as seen in Figure \ref{fig:rnn_prediction_sw_arch} that communicate with each other using the Open Sound Control (OSC) network protocol \cite{Wright1997}. This allows the interaction and visualisation frame-rate to be independent of the sequence generation frame-rate. It also allows us to run the Server and Visualiser on different (networked) computers if need be (e.g. a powerful GPU-based server for running the models, and a less powerful front-end computer for visualisation and interaction).

\section{Results and discussion}
In this study we train an ensemble of LSTM RNNs, with each model trained on a different corpus, and we build an interactive prediction and visualisation system which mixes each model's predicted probability distributions via mixture weights, controlled in real-time via a user's gestures.

The system works as desired and allows users to continuously `steer' the output while text is being generated, seamlessly morphing between styles, in effect `conducting' the generation of text. Figure \ref{fig:screen} shows an example output.

We also observe some interesting behaviour. When multiple models are active with roughly equal mixture weights, and the system is fed a sequence containing words or phrases that are common to all models, the probabilities for the common characters accumulate whilst probabilities specific to individual models are suppressed, i.e. when multiple models are active the system tends towards common words and phrases. 

Sometimes, while a sequence is being generated, a particular model might output a spiking probability distribution (i.e. very high confidence for a particular character). If at that point other models output wider distributions (i.e. lower confidence aross multiple characters), then the first model will overpower and dominate the sequence generation.

E.g. If at any time-step the input sequence ends with `$the \ house \ $', the \textit{Bible} model predicts the letter $o$ with very high confidence (to eventually lead on to `$the \ house \ of \ [Judah/Jeremiah/Noah/Isaac/etc...]$'). Even if at that time-step, the \textit{Bible} model has a lower mixture weight than the other models, it is probable that it might overpower the other models' probability distributions and cause the $o$ to be dominant in the final joint probability distribution. This is quite likely to start a positive feedback loop and that model will stay in control of the sequence generation until it reaches a point where its probability distribution widens, and another model spikes. So it's very possible to see hints of love songs, philosophy or poetry within C comments and variable names or \LaTeX{} equations. It seems there are `hand-over' words or sequences which are common to many models, but have stronger connotations in some models over others. This is of course further guided by the user's actions, who can choose to push further towards the emerging theme, or pull towards another style and seamlessly go from one style to another over these hand-over words.

It is also worth noting, that like most character based LSTM models, the output is quite nonsensical. The only long-term dependencies which are preserved are in formatting and syntax, and there is only meaning within the space of a few neighbouring words. Nevertheless, it's still very interesting to see the model produce words and phrases very much in the style of the associated texts, with correct formatting, punctuation, indentation etc. There are also some nonsensical words, spelling mistakes and incorrect punctuation. As well as being due to mixing probability distributions, this behaviour is also observed in \textit{single} models, and is most likely due to the relatively small training set (a few hundred KB for some) and `unclean' data. With more time dedicated to collecting more training data and cleaning it, this is likely to be improved.

\subsection{Future work}
Mixing models with approximate equal weights generally works when the number of models is low (e.g. $n < 4$). When we go beyond that, the sequence occasionally diverges away from comprehensible words, towards what appears to be random sequences of characters. This is accentuated by the fact that the system is predicting on a character-by-character level with no foresight beyond that. In order to overcome this problem, we are planning on implementing a beam search \cite{Graves2012} with limited depth, whereby we sample multiple times per time-step, and explore (i.e. resample) each sample a few time-steps into the future, scoring each path on the sum of the log-probabilities accumulated along the way, then pruning and selecting accordingly.

As opposed to training many models independently on different corpora, another approach we are looking at is using a \textit{single} model trained on the entire corpora. We would then look to control the output via manipulating the internal state of the LSTM. This has advantages and disadvantages, particularly when it comes to adding a new corpus (i.e. 'style') to the system.

With our current approach, adding a new 'style' is relatively quick, since we only need to train a new model on just the new corpus. However scalability becomes an issue during \textit{deployment}. Since all of the models are run during prediction, having too many models can be a bottleneck. We have implemented an optimization such that if the mixture weight of a model is less than 5\% at a particular time-step, we don't run the model. This allows us to have many (10+) models loaded in the system and still retain real-time performance if not all models are mixed in at every time-step. However, as we dynamically mix in more models, the rate of sequence generation drops from around 10-20 char/second (for 2-4 models) to 1 char/second (8+ models) on a high-end laptop. N.B. Since the Visualiser is a separate process to the Server running the models, the Visualiser framerate is always real-time at 60fps, so the response of the interactivity and visualisation doesn't suffer, but characters are output at a slower rate.

With a single monolithic model, prediction performance is less of an issue, since we will always be running a single model. However, in this case \textit{training} performance can become an issue. To add a new 'style', we will have to incrementally train the model on the new corpus, while makig sure it maintains prediction accuracy on the previous collection of corpora. As we add more and more styles, this is likely to have a big impact on training times and memory requirements. 

Finally, we are currently working with character based text models because the dimensions are relatively low and discrete, training data is easily accessible and judging the outputs is quite straightforward. However we are planning on applying these techniques to higher dimensional and continuous domains such as music, sound and vector graphics. 

\section{Acknowledgements}
In addition to my ongoing research in this field as part of my PhD, this work was further supported by a placement at Google's Artists and Machine Intelligence Program. In that capacity I'd like to especially thank Mike Tyka, Kenric McDowell, Blaise Aguera y Arcas and Andrea Held for the organization, inspiration and support; and Doug Fritz, Douglas Eck, Jason Yosinki, Ross Goodwin, Hubert Eichner and John Platt for the inspiring conversations and ideas.

The Server is implemented with \textit{Keras} \cite{Keras}, using the \textit{Theano} \cite{Theano} backend, while the Visualiser is implemented with \textit{openFrameworks}, a C++ framework for creative development \cite{openframeworks}. This work wouldn't have been possible without these wonderful opensource toolkits. 

\small
\bibliographystyle{iccc}
\bibliography{msa}

\begin{thebibliography}{10}

\bibitem{bengio1994learning}
Y.~Bengio, P.~Simard, and P.~Frasconi.
\newblock {Learning long-term dependencies with gradient descent is difficult}.
\newblock {\em IEEE Transactions on Neural Networks}, 5(2):157--166, 1994.

\bibitem{Boulanger-Lewandowski2012}
N.~Boulanger-Lewandowski, P.~Vincent, and Y.~Bengio.
\newblock {Modeling Temporal Dependencies in High-Dimensional Sequences:
  Application to Polyphonic Music Generation and Transcription}.
\newblock {\em arXiv preprint arXiv:1206.6392}, 2012.

\bibitem{Keras}
F.~Chollet.
\newblock {Keras: Deep Learning library for Theano and TensorFlow}, 2015.

\bibitem{friis2016}
L.~Crnkovic-friis and L.~Crnkovic-friis.
\newblock {Generative Choreography using Deep Learning}.
\newblock {\em arXiv preprint arXiv:1605.06921}, 2016.

\bibitem{Dietterich2000}
T.~G. Dietterich.
\newblock {Ensemble Methods in Machine Learning}.
\newblock {\em MCS '00: Proceedings of the First International Workshop on
  Multiple Classifier Systems}, pages 1--15, 2000.

\bibitem{Eck2002}
D.~Eck and J.~Schmidhuber.
\newblock {A First Look at Music Composition using LSTM Recurrent Neural
  Networks}.
\newblock {\em Istituto Dalle Molle Di Studi Sull Intelligenza Artificiale},
  103, 2002.

\bibitem{Gers2000}
F.~A. Gers and J.~Schmidhuber.
\newblock {Recurrent nets that time and count}.
\newblock In {\em Neural Networks, 2000. IJCNN 2000, Proceedings of the
  IEEE-INNS-ENNS International Joint Conference on}, pages 189--194. IEEE,
  2000.

\bibitem{goodwin-wordsynth}
R.~Goodwin.
\newblock {Word Synth}, 2016.

\bibitem{Graves2012}
A.~Graves.
\newblock {Sequence transduction with recurrent neural networks}.
\newblock {\em arXiv preprint arXiv:1211.3711}, 2012.

\bibitem{Graves2013}
A.~Graves.
\newblock {Generating sequences with Recurrent Neural Networks}.
\newblock {\em arXiv preprint arXiv:1308.0850}, 2013.

\bibitem{Graves2009}
A.~Graves, M.~Liwicki, S.~Fern{\'{a}}ndez, R.~Bertolami, H.~Bunke, and
  J.~Schmidhuber.
\newblock {A novel connectionist system for unconstrained handwriting
  recognition}.
\newblock {\em IEEE Transactions on Pattern Analysis and Machine Intelligence},
  31(5):855--868, 2009.

\bibitem{Greff2015}
K.~Greff, R.~K. Srivastava, J.~Koutn{\'{i}}k, B.~R. Steunebrink, and
  J.~Schmidhuber.
\newblock {LSTM: A Search Space Odyssey}.
\newblock {\em arXiv preprint arXiv:1503.04069}, page~10, 2015.

\bibitem{Gregor2015}
K.~Gregor, I.~Danihelka, A.~Graves, and D.~Wierstra.
\newblock {DRAW: A Recurrent Neural Network For Image Generation}.
\newblock {\em arXiv preprint arXiv:1502.04623}, 2015.

\bibitem{Hinton2012}
G.~Hinton, L.~Deng, D.~Yu, G.~E. Dahl, A.-r. Mohamed, N.~Jaitly, A.~Senior,
  V.~Vanhoucke, P.~Nguyen, T.~N. Sainath, and B.~Kingsbury.
\newblock {Deep Neural Networks for Acoustic Modeling in Speech Recognition}.
\newblock {\em IEEE Signal Processing Magazine}, 29(6):82--97, 2012.

\bibitem{Hochreiter1991}
S.~Hochreiter.
\newblock {\em {Untersuchungen zu dynamischen neuronalen Netzen}}.
\newblock PhD thesis, Technische Universit{\"{a}}t M{\"{u}}nchen, 1991.

\bibitem{Hochreiter1997}
S.~Hochreiter and J.~Schmidhuber.
\newblock {Long Short-Term Memory}.
\newblock {\em Neural Computation}, 9(8):1735--1780, 1997.

\bibitem{Karpathy2015a}
A.~Karpathy.
\newblock {The Unreasonable Effectiveness of Recurrent Neural Networks}, 2015.

\bibitem{openframeworks}
Z.~Lieberman, T.~Watson, and A.~Castro.
\newblock {OpenFrameworks}, 2016.

\bibitem{Nayebi2015}
A.~Nayebi and M.~Vitelli.
\newblock {GRUV : Algorithmic Music Generation using Recurrent Neural
  Networks}.
\newblock 2015.

\bibitem{Pham2014}
V.~Pham, T.~Bluche, C.~Kermorvant, and J.~Louradour.
\newblock {Dropout Improves Recurrent Neural Networks for Handwriting
  Recognition}.
\newblock In {\em Frontiers in Handwriting Recognition (ICFHR), 2014 14th
  International Conference on}, pages 285--290. IEEE, 2014.

\bibitem{Schmidhuber2015}
J.~Schmidhuber.
\newblock {Deep Learning in Neural Networks: An Overview}.
\newblock {\em Neural Networks}, 61:85--117, 2015.

\bibitem{Sturm2015}
B.~Sturm.
\newblock {Recurrent Neural Networks for Folk Music Generation}, 2015.

\bibitem{Sutskever2013}
I.~Sutskever.
\newblock {\em {Training Recurrent neural Networks}}.
\newblock PhD thesis, University of Toronto, 2013.

\bibitem{Martens2011}
I.~Sutskever, J.~Martens, and G.~Hinton.
\newblock {Generating Text with Recurrent Neural Networks}.
\newblock In {\em Proceedings of the 28th International Conference on Machine
  Learning (ICML-11)}, pages 1017--1024, 2011.

\bibitem{SutskeverVinyals2014}
I.~Sutskever, O.~Vinyals, and Q.~V. Le.
\newblock {Sequence to Sequence Learning with Neural Networks}.
\newblock In {\em Advances in Neural Information Processing Systems (NIPS)},
  pages 3104--3112, 2014.

\bibitem{Theano}
{The Theano Development Team}.
\newblock {Theano: A Python framework for fast computation of mathematical
  expressions}, 2016.

\bibitem{Wright1997}
M.~Wright and A.~Freed.
\newblock {Open Sound Control: A new protocol for communicating with sound
  synthesizers}.
\newblock In {\em Proceedings of the 1997 International Computer Music
  Conference (ICMC)}, 1997.

\bibitem{Wu2016}
Z.~Wu and S.~King.
\newblock {Investigating Gated Recurrent Neural Networks for Speech Synthesis}.
\newblock In {\em IEEE International Conference on Acoustics, Speech and Signal
  Processing (ICASSP)}, pages 5140--5144. IEEE, 2016.

\bibitem{Zaremba2014}
W.~Zaremba and I.~Sutskever.
\newblock {Learning to Execute}.
\newblock In {\em 2nd International Conference on Learning Representations
  (ICLR2014)}, 2014.

\end{thebibliography}

\clearpage
\appendix
\section{Sample output}
Sample output, interactively guided by the user in realtime. 
\scriptsize
\begin{verbatim}
But that's okay because I'm here right now to thank you in person and that was my term and I_m not going
to be a real wall. That_s what happened in the world. We don_t know what they do we don_t.
And then I like the money. I will make no man an ass unto the wicked.

7:15 For they shall see the prophets, they shall come and scatter them in the way, and their ear shall be confounded.

11:22 And the LORD spake unto Moses, unto the priests and of the conditions of
Lemma \ref{lemma-finite-colimit-finite-locally-free}.
\end{proof}

\begin{lemma}
\label{lemma-dimension-ring-property-local}
Let $f : X \to Y$ be a morphism of schemes.
Let $X$ be a self to be supposed to have made absolute being, and not of the meaning of the consider of the similar
discussion and the latter of the latter and I say all I_m for the way

All of my side
I can see you to think I_ve done
It_s all I really do
I can see your love to love the best that I mean
I should have got to fall in love,
I can see your rain
I_m always between the Iraqi Committee of Security Council on the military officers to the UK statement of the US
and the public of the court that the wart and her breast,
The winding soul of the sunny hour
Where some more lady had seems to sound and steal
In the secret song.

The flight and season'd Folly a double angels, and sent down to you when they are unjust?
When he had commanded the angels and his people who do not believe in His creatures.
It is He who created the heavens and exclusive under-entity and callbacks and all the
     * times which can be not an interrition of death: and he shall take away the day could
     * be registered and return the timer completely can be changed
     * only and we can be released by call_rcu_tasks() that is returned
     * for the task is not exiting the same partition descriptor.
     */
    raw_spin_lock_irqsave(&cpu_ba_flip_lock, flags);
    if (!ret)
        goto out_unlock;

    return ret;
}

static inline void __init thread_create(veronest and his way to the ship */
{
    int i;

    if the servants of the town and the servants of the ship we will say. When I saw this that he was returned
 * that is dead to the time that has an the timer t to with the mercy); and those who deny the straight path.
They shall be your father without about the dead from your Lorded containing/state
 * and one warning should have intconsider the construction of the products
of sheaves of sets on $\mathcal{C}$, seeing the king of Babylon have said "Well, we_re leading a lot of control.
/
    if (!param) {
        /* Yeah
Sometimes it comes to be through
I believe and it_s over now.
I fully the teacher is the patient's phenomena and the dream which do not waking in the same time that is really a similar
symbol, so to individual and possible policy of the basis of Iraq and Jerusalem to be heavy with thee.

10:17 Then said Joseph, where $X$ is a finite dimensional ring of the dark
I_m not there to fall with the head of the same thing as a distinct and a call not make A
\end{verbatim}
\end{document}